\documentclass[]{scrartcl} 

\usepackage{url,hyperref,lineno,microtype,subcaption}
\usepackage[onehalfspacing]{setspace}

\usepackage{array, multirow, bigdelim, makecell, booktabs} 
\usepackage{graphicx}
\graphicspath{{images_forupload/}}
\usepackage{algorithmic,algorithm}
\usepackage{siunitx}[=v2]
\usepackage{graphicx,import}
\usepackage{hyperref}
\usepackage{forest}
\usepackage{bm}
\usepackage{amsmath,amssymb,amsthm}
\usepackage{textcomp}
\usepackage[backend=biber,%
                      bibencoding=utf8,%
                      citestyle=numeric,%
                      giveninits=true,%
                      isbn=false,%
                      sortcites=true,%
                      natbib=true]{biblatex}
\providecommand{\keywords}[1]{\textbf{\textit{Keywords ---}} #1}
\author{Philipp Rodegast\thanks{Institute of Engineering and Computational Mechanics, University of Stuttgart, Pfaffenwaldring 9, 70569 Stuttgart, Germany. (\url{philipp.rodegast,jonas.kneifl,joerg.fehr@itm.uni-stuttgart.de})}
\and Steffen Maier\footnotemark[1] 
\and Jonas Kneifl\footnotemark[1] 
\and J\"org Fehr\footnotemark[1] }
\title{On using Machine Learning Algorithms for Motorcycle Collision Detection}

\begin{document}


\maketitle

\textbf{Abstract} 
Globally, motorcycles attract vast and varied users. However, since the rate of severe injury and fatality in motorcycle accidents far exceeds passenger car accidents, efforts have been directed toward increasing passive safety systems. Impact simulations show that the risk of severe injury or death in the event of a motorcycle-to-car impact can be greatly reduced if the motorcycle is equipped with passive safety measures such as airbags and seat belts. For the passive safety systems to be activated, a collision must be detected within milliseconds for a wide variety of impact configurations, but under no circumstances may it be falsely triggered. For the challenge of reliably detecting impending collisions, this paper presents an investigation towards the applicability of machine learning algorithms. First, a series of simulations of accidents and driving operation is introduced to collect data to train machine learning classification models. Their performance is henceforth assessed and compared via multiple representative and application-oriented criteria.


\section{Motivation}
\label{sec:einleitung}

Globally, powered two-wheelers, more commonly called 'motorcycles,' attract vast and varied users~(\cite{Haworth12}). Dependent on the region, they enjoy popularity as~(i)~a compact means of individual transportation in increasingly congested urban traffic and tight parking conditions,~(ii)~cheap individual mobility due to a lower price, lower energy consumption, and lower maintenance costs compared to, e.g., passenger cars, and~(iii)~a leisure or exciting sports activity due to its unique handling experience with the feeling of freedom when riding. However, their poor passive safety poses an excessive risk in road traffic at considerable social costs. 

The risk of suffering severe injuries or dying in an accident is more than twice as high for motorcyclists than for occupants of cars. Worldwide, about~375,000 drivers and passengers of two- or three-wheeled vehicles die each year, according to~\cite{who2018}. In~2021,~302~(2.2\%) motorcyclists killed and~5,230~(38.2\%) motorcyclists seriously injured were recorded in~13,702 motorcycle accidents with personal injury in Germany alone~(\cite{Destatis22}). In that same period, in~160,771 car accidents involving injuries to persons, only~1,433~(0.9\%) were killed, and~30,902~(19.2\%) severely injured car drivers were registered. The higher risk of injury and fatality rate is attributed to the lack of passive safety systems. While passenger cars benefit from extensive safety measures such as airbags and seatbelts in an enclosed passenger cell with an energy-absorbing crumple zone, motorcycles lack these features. There are many different active safety systems that improve the active safety of motorcycles~(\cite{SavinoEtAl20} including motorcycle-detecting automatic emergency braking which is integrated into passenger vehicles cars~(\cite{DeanEtAl21})). Studies show that they can potentially prevent a significant proportion of accidents, but they are not able to avoid all collisions in the future~(\cite{SavinoPieriniFitzharris19,DeanEtAl21}). Instead, nowadays motorcyclists rely on personal protective equipment such as helmets, safety clothing, and protectors, which, as the above statistics of accident outcomes show, have poor protective performance.

As part of a research project, the Institute~of~Engineering~and~Computational~Mechanics~(ITM) of the University of Stuttgart is investigating a new passive safety concept for motorcycles. The concept includes, among other features, seatbelts that restrain the rider to the vehicle and airbags in order to decelerate the rider in a controlled manner. The strategy for an impact scenario is no longer to detach the rider from the motorcycle so that, in the best-case scenario, to fly over the accident opponent without violent contact but to restrain and decelerate the rider with airbags and belts. Crash simulations of a set of frequent impact configurations show that the likelihood of injuries is significantly reduced by applying the novel safety concept~(\cite{Maier2020,MaierEtAl21,MaierEtAl22}). Like other future deployable passive safety systems for PTWs~(e.g., airbags for scooters from~\cite{Autoliv22}), the concept depends on reliably detecting a hazardous impact opponent with the need for a sufficiently short decisional delay.


Available since 2007 with a front airbag, the Honda Goldwing GL1800 heavy touring motorcycle is the only motorcycle available on the market that detects an impact with an accident opponent to activate passive safety systems. However, because it is designed for a frontal impact, the sensor system in the front fork does not detect side or rear impacts~(\cite{KobayashiMakabe13,kuroe2005}). In~\cite{engel1992}, a whole range of other sensors are proposed to detect a motorcycle impact as soon as possible. A study on the proposed safety concept~(\cite{Daub21}) predicts that a large number of such sensors is needed to detect many accident configurations reliably without unwanted faulty activation in expected driving conditions. As a result, the greater the number of sensors, the more complex and difficult it is to optimize and validate the decision-making process for activation. Ultimately, this is a very complex task for conventional methods such as threshold-driven detection logic and has strict requirements on validation and verification, see, e.g.,~\cite{Leschke20}. Hence, this work is dedicated to the search for other methods capable of reliably detecting accidents within milliseconds. Algorithms from the field of machine learning are promising candidates for this task as, on the one hand, they are able to detect non-obvious patterns in data, which are therefore ignored in classical detection logic. On the other hand, many of them are very efficient and can therefore be quickly evaluated and used on hardware with limited computing power, such as found in the control units of a motorcycle.

Utilization of machine learning methods for crash predictions is already being investigated in a wide range of applications. Among others, many publications deal with automated real-time traffic supervision, which aims towards identifying potentially hazardous road sections. In~\cite{theofilatos2019}, different deep learning and machine learning techniques are compared to predict real-time crash occurrence based on traffic data as well as weather information. Similar efforts are taken for crash detection as well as risk estimation~(\cite{Huang2020}) and identification~(\cite{Yang2022}) on freeways which could be used for traffic management. When it comes to crash predictions for motorcycles, on the contrary, the number of research projects and publications is lower. One field of research where machine learning~(ML) methods are useful is the investigation on crash- or injury-severity in motorcycle accidents, which gives useful insights into how motorcycle riding can be rendered less dangerous~(\cite{rezapour2020}). Another general recurring approach for cars is to use smartphone-based crash detection~(\cite{Aloul2015, Condro2012, mian2021, deng2021}), where the smartphone serves as a measurement tool as well as an automated emergency notification system. Instead of using smartphone-based measurements, \cite{Choi2021} use an ensemble-based crash detection on data emerging from dashboard cameras. 

However, the possibility of using the motorcycle sensors themselves is often overlooked since most publications do not consider safety frameworks for motorcycles. Consequently, decisions are based on a few non-specialized measurements so that the detection performance is highly limited, also allowing for false positive detection. Furthermore, most publications rely on difficult-to-access and poorly available real-world data where it is hard to reproduce exactly those scenarios crucial for accident detection. As we use simulation models to generate the training data, we are able to generate specific and well-defined parametric scenarios describing a wide variety of typical accidents and non-critical driving scenes. To the authors' best knowledge, no research currently investigates the applicability of machine learning methods as a real-time, onboard crash prediction mechanism in order to trigger a given passive safety protocol.

This leads to the fundamental research question of this work: Is a ML model capable of accurately identifying an impact in due time? Responding to this question, this paper deals first with producing an expressive database that is consequently used to train a selection of~ML models and to evaluate their performance. The following describes the used simulation model, illustrates how the database is produced and trains a selected set of ML classification models on this data.



\section{Materials and Methods}
\label{chap:method}

First, the simulation model is described, which is used to collect both operational driving data (regular use) and collision data (accidents). The collision data exclusively includes impacts into a car as a collision partner. It does not include solo accidents, since this study merely serves as a proof of concept and the simulation model is reused from a preceding study~(\cite{Maier2020}). Subsequently, data generation and processing are addressed. In the last section, ML methods are implemented and optimized.

\subsection{Simulation Model}

Data collection is accomplished solely by means of simulation as it offers the opportunity to cheaply evaluate and measure various operational driving and collision scenarios. In addition, the motorcycle concept currently only exists virtually, and real-world driving and crash tests are costly and pose great challenges to represent targeted scenarios. The simulation model is implemented in Siemens Simcenter's Madymo~(version MADYMO~2020.1), which serves as a multibody~(MB) finite-element~(FE) crash-simulation environment. Madymo is a simulation environment for physical systems with a focus on vehicle collision dynamics and passenger safety and injury assessment. It combines MB system capabilities for large rigid body motions as well as FE analysis for structural behavior. Here, simulating crash dynamics not only saves costs but can also serve to reduce time to market significantly.

\subsubsection{Motorcycle Model And Accident Opponent}

\begin{figure}
  \begin{center}   
	\includegraphics[width=8.5cm]{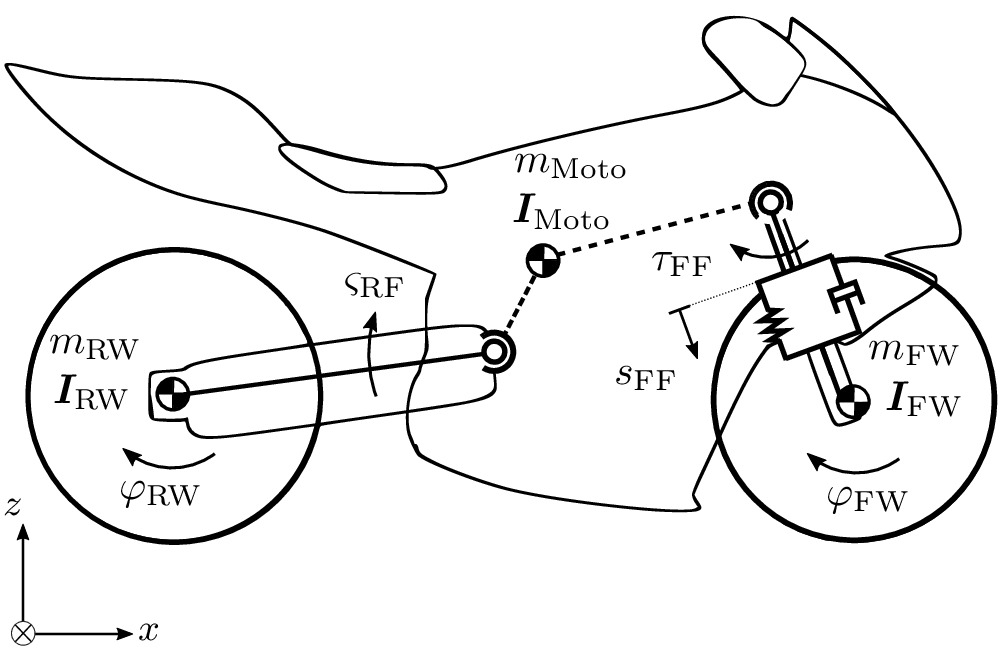}     
	\caption{Three mass rigid body model of the motorcycle: A frame (Moto) and two wheels (FW, RW).}
	\label{fig:MC}
  \end{center}
\end{figure}

The motorcycle is modeled with three masses, see~Fig.~\ref{fig:MC}, defined by their mass~$m$, rotational inertia~$\bm{I}$, and geometry. The bodies are linked via kinematic joints. The motorcycle chassis is attached to the suspended front fork~(FF) and rear swing~(RS). The telescopic front fork translates linearly~($s_{\mathrm{FF}}$), whereas the rear swing rotates angularly~($\varsigma_{\mathrm{FF}}$). Both suspensions attach to their wheels~(FW,~RW) via hubs, around which these rotate~($\varphi_{\mathrm{FW}}, \varphi_{\mathrm{RW}}$). To model structural deformation in an impact, the front fork allows for angular deformation~$\tau_{\mathrm{FF}}$. It depicts a rotation of the front fork inwards around the steering hub. As an impact opponent vehicle, the simulation setup comprises a collision partner, which is represented by a~1987 Ford Scorpio configured according to \cite{hiemer2005}. The model is part of a broader modeling and simulation strategy ranging from a multibody-system~(\cite{Maier2020}), a coupled multibody- and finite-element setup~(\cite{MaierEtAl21,MaierEtAl21b}), to a full finite element model~(\cite{MaierFehr23}). The multibody model used here is chosen in order to simulate long scenarios that span many seconds. For a thorough overview of the models that also include the passive safety systems, be referred to~\cite{Maier2020, MaierFehr23b}.

In contrast to investigations of passive safety measures in the course of an accident, which are already well-established, the period of interest for this study includes only a few moments after the impact. This means that the period of interest for this study occurs well before the passenger comes in contact with the airbags or is significantly restrained by the belts allowing to reduce the overall computational costs significantly. On the one hand, this allows stopping the crash simulations before any computationally complex passenger-safety-system interactions occur. On the other hand, the passive safety system, i.e., thigh belts and airbags, and the rider model can be excluded from the simulation setup as they have no qualitative influence on the simulation results in the period under consideration. These simulation setup modifications are beneficial for the overall numeric costs and allow for the execution of a far greater number of simulations than what would have been feasible within the same time with the inclusion of rider and passive safety systems. 

A crucial issue with generating data from the model is that, due to complexity, the model does not allow for the application of lateral dynamics. Cornering behavior can thus not be incorporated into the dataset. Hence, only longitudinal dynamics are covered by the training data. This circumstance has far-reaching ramifications: Since, for example, the steering angle moves out of the $0^\circ$ position in case of a head-on crash and by design never leaves this position during normal~(non-crash) riding, the classifier would be induced false knowledge and would most likely decide solely on the basis of the steering angle signal. To circumvent this contingency, signals that contain information about lateral motion are to be strictly neglected in order not to overestimate the decision-making ability of a classification model. This poses a major restriction on parts of the available sensor data, as certain DOFs are only changed when the motorcycle collides with an opponent. Signals that are affected by the restriction are the motorcycle body's velocity and acceleration, both linear and angular.

In Table~\ref{tab:signals}, all modelled sensors and available signals are listed. The first five signals are subject to dimensional limitation. The used component of these signals is given in the last column. In order to emulate the response of a tire pressure sensor, the resulting contact force between the crash opponent and each tire is combined with the contact force between each tire and the road surface, yielding the residual contact forces~$f_{FW}$ and~$f_{RW}$ for the front and respectively the rear wheel.

\begin{table}
  \caption{Recorded output signals from the motorcycle simulation model. }
  \label{tab:signals}
  \begin{center}
    \begin{tabular}{p{3cm} p{7cm} s[per-mode=symbol,table-unit-alignment=left] r p{1cm}}
		\hline
		abbreviation  &  description & unit & comp.\\ \hline
		Body lin. vel.			& motorcycle body linear velocity & \m\per\s & $x$\\ 
		Body lin. acc. 		& motorcycle body linear acceleration& \m\per\s\squared& $x$ \\
		Body ang. vel. 		& motorcycle body angular velocity &\radian\per\s&$y$\\	
		Body ang. acc.  		& motorcycle body angular acceleration&\radian\per\s\squared& $y$\\	
		FW ang. vel.  		&  front wheel angular velocity & \radian\per\s \\	
		FW ang. acc.  	&  front wheel angular acceleration &\radian\per\s\squared\\	
		RW ang. vel.  		& rear wheel angular velocity &\radian\per\s \\	
		RW ang. acc.   	&  rear wheel angular acceleration &\radian\per\s\squared\\
		RS ang. pos.  			& rear swing angular position &\radian \\
		RS ang. vel. 	& rear swing angular velocity &\radian\per\s \\	
		RS ang. acc.  	& rear swing angular acceleration & \radian\per\s\squared \\
		FS lin. pos.				& front suspension linear position &\m \\
		FS lin. vel.  			& front suspension linear velocity &\m\per\s \\	
		FS lin. acc.  			&front suspension linear acceleration &\m\per\s\squared \\
		FD ang. pos.				& front deflection angular position & \radian\\
		FD ang. vel. 		& front deflection angular velocity & \radian\per\s\\	
		FD ang. acc.   		& front deflection angular acceleration & \radian\per\s\squared \\
		FW cnt. force					& front wheel contact force &\N \\	
		RW cnt. force 					& rear wheel contact force &\N\\	
		FW/RW vel. diff. 			&wheel speed differential &\radian\per\s\\
		FW/RW acc. diff. 			&wheel acceleration differential & \radian\per\s\squared \\
		cnt. sensor left 				& contact sensor left & boolean\\
		cnt. sensor right 				& contact sensor right & boolean\\
		\hline
    \end{tabular}
  \end{center}
\end{table}

\subsection{Data Acquisition}

A notable benefit of the proposed method is that the data used to train ML models does not come from logged real-world sensor data but from closely monitored simulations. This means that each individual sample can be assigned to a scenario and, therefore, also state (non-crash/crash). The knowledge about the state introduces the ability to use supervised learning methods. A switch is incorporated into the model in order to automate the labeling process. It is flipped as soon as a part of the motorcycle comes in contact with the car. The switch is allowed only one initial flip since, for the purpose of this elaboration, a crash does not stop until the simulation terminates. The switch reliably distinguishes between normal non-crash operation and crash scenarios, and is, thus, eligible to be further used as a class label for the machine learning classification application. 

A fundamental operation of this investigation is to produce data covering the whole spectrum of both non-crash driving and crash scenarios. Training data must contain all necessary information to reliably differentiate between those two states, but it can only be composed of a multitude of individual simulations. 
The fact that the labeling process is not a task to be carried out manually but rather by the simulation itself enables to consider all necessary scenarios. Henceforth, a scenario-parameter-based simulation definition is used to fully exploit the fact that individual simulation postprocessing is not required. Consequently, a set of scenario parameters with an allocated range defines each subset. The individual subsets are then merged to form a comprehensive database.

In order for the parameter space to be sufficiently covered by a given amount of instances, latin-hypercube sampling~(LHS) is applied.~LHS, in contrast to simple-random-sampling~(SRS), subdivides each parameter's range into equal-sized subdivisions, thus effectively partitioning the entire parameter space into hypercubes, called \textit{strata}. A random sample is generated from each \textit{stratum}, thus ensuring that the resulting distribution is fully representative of a given population~(\cite{mckay1979}). All resulting simulative measurements are available under~\cite{darus-3301}.

\subsubsection{Class \textit{A}: Uncritical Data}
Within the class of uncritical data, simulations are divided into two subsets. The first of which uses a sinusoidal base structure as a road profile. The second subset supplementary adds specified use cases to the database that can not be reproduced by the sinusoidal subset.

\subsection*{Scenario Set \textit{A.1}}

\begin{figure}
  \begin{center}
	\includegraphics[width=8.5cm]{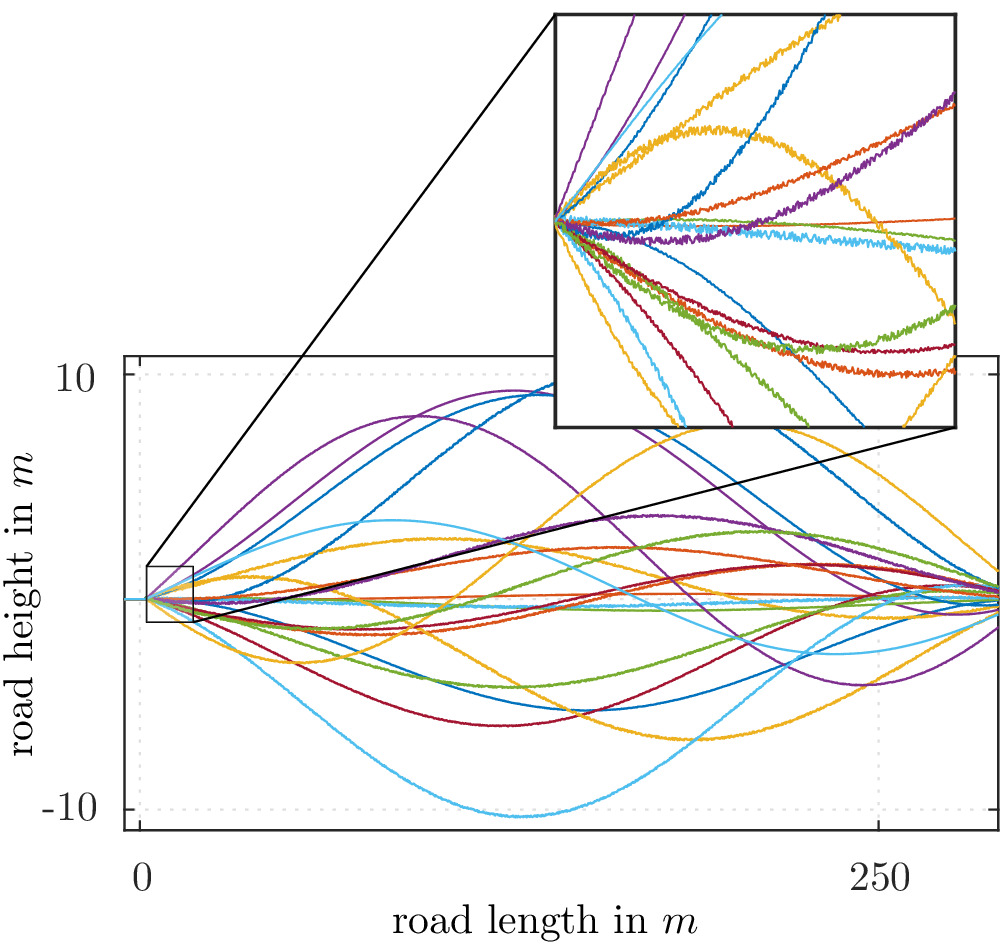}        
  \end{center}
  \caption{Road profiles for non-crash scenario set~\textit{A.1}. The road consists of a start platform which is connected to a randomly generated sinusoidal profile with a maximum gradient of up to 12\% and a superimposed noise.}
  \label{fig:road}
\end{figure}

A mesh of rigid shell elements which is created before each simulation serves as contact surface for the motorcycle. This road model has a total length of~\SI[]{300}[]{\metre} and a segment length of~\SI[]{0.2}[]{\metre}. The sine amplitude ranges from~\SI[]{0}[]{\metre} to~\SI[]{6}[]{\metre}. The first subset's road profile is described by a sinusoidal base profile with added noise in order to mimic road-unevenness. By ensuring that the interval length corresponds to the road length, the steepest section is limited to a road gradient of~12\% which amounts to the maximum gradient found on open roads according to~\cite{liu2017}. Phase shift and noise amplitude are also incorporated as parameters for additional variation. A set of~20 exemplary road profiles is depicted in Fig.~\ref{fig:road}. The figure additionally depicts an expanded view in order to illustrate the superimposed noise. The motorcycle's initial velocity ranges from~\SI[]{3}[]{\metre\per\s} up to~\SI[]{23}[]{\metre\per\s}.  The speed range is chosen to correspond to that of the crash scenarios (\ref*{fig:b3}). Otherwise, if speed distribution in one class far exceeds that of the other class, the classification model would tend to incorporate that bias. Lastly, brake and acceleration torque acting on the wheel hubs are derived from one single parameter~(since they cannot occur at once).~\cite{corno2008} show that the mean braking torque of a motorcycle can be assumed to be~500\,Nm and is distributed, so that~70\% is directed to the front wheel and~30\% to the rear wheel. Likewise,~\cite{hauser2006} investigate the acceleration of high-performance race motorcycles. The authors state that the braking torque at the wheel can reach up to~1000\,Nm. As a conservative estimate the acceleration torque limit is set to~500\,Nm since the motorcycle under investigation here is not designed for racing. Scenario set~\textit{A.1} is composed of~100 entities that are parametrized via LHS. The parameters with their corresponding variation range are listed in Tab.~\ref{tab:a1}.

\begin{table}
  \caption{Parameter space for non-crash scenario set~\textit{A.1}. }
  \label{tab:a1}
  \begin{center}
	\begin{tabular}{p{3cm} r c r s[per-mode=symbol,table-unit-alignment = left]}
	\hline
		parameter      & \multicolumn{3}{c}{range} & unit\\ \hline
		sine amplitude & 0 &:& 6& \si{\m} \\
		phase shift    & 0 &:&  360 & \si{\degree} \\
		road 'noise'   & 0 &:& 0.025& \si{\m} \\
		$v_{\mathrm{Moto,init,A1}}$ & -1.2 &:& 1.2& \si{\meter/\second} \\
		$t_{\mathrm{accel/decel,A1}}$ & -500 &:& 500& \si{\N\m} \\
		\hline
	\end{tabular}
  \end{center}
\end{table}

\subsection*{Scenario Set \textit{A.2}}

\begin{figure}
  \begin{center}
	\includegraphics[width=\textwidth]{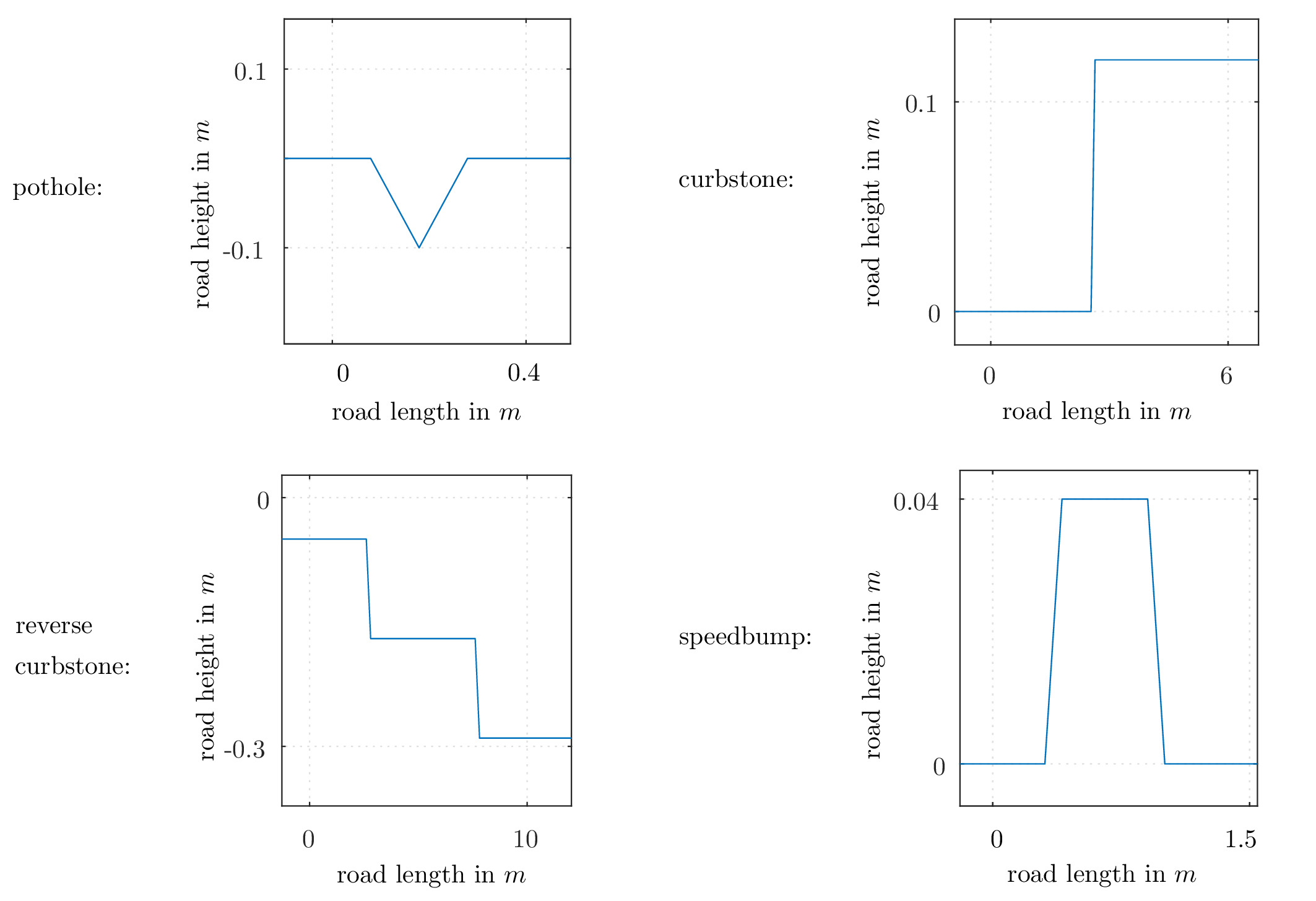}        
  \end{center}
  \caption{Additional road profiles for non-crash scenario set~\textit{A.2}.}
  \label{fig:a2}
\end{figure}

Although the parametrized scenario generation already covers a wide range of applications, some use cases can not be emulated by this method and should still be considered in the database. These scenarios, referred to as set~\textit{A.2}, are designed manually and appended onto the dataset. Namely, these simulations are intended to replicate the following use cases:~(i)~riding over potholes,~(ii)~approaching a curbstone,~(iii)~riding down multiple curbstones, and~(iv)~riding over speedbumps. Road profiles that are designed to imitate these obstacles are depicted in Fig.~\ref{fig:a2}. In combination with initial velocities lying within the same range as in subset~\textit{A.1}~, see Tab.~\ref{tab:a1} and using LHS as sampling method, a total of~20 simulations~(4~in each use case) are performed to build subset~\textit{A.2}.

\subsubsection{Class \textit{B}: Crash Data}

The data containing accidents involves more than just head-on collisions, where simple threshold-based decision logic would suffice for detection. Instead, the aim is to examine a broad range of conceivable impact scenarios. Consequently, parametrized scenario generation is resorted to again. Two different base architectures are designed to cover the motorcycle striking a stationary car at different angles and velocities~(subset~\textit{B.1}) and a car striking the stationary motorcycle at different angles and velocities~(subset~\textit{B.2}). Additionally, a set of ISO~13232 crash scenarios are simulated and preserved in order to validate the classifier's ability to generalize after being trained.  

\subsection*{Scenario Set \textit{B.1}}

\begin{figure}
  \begin{center}  
		\includegraphics[width=8.5cm]{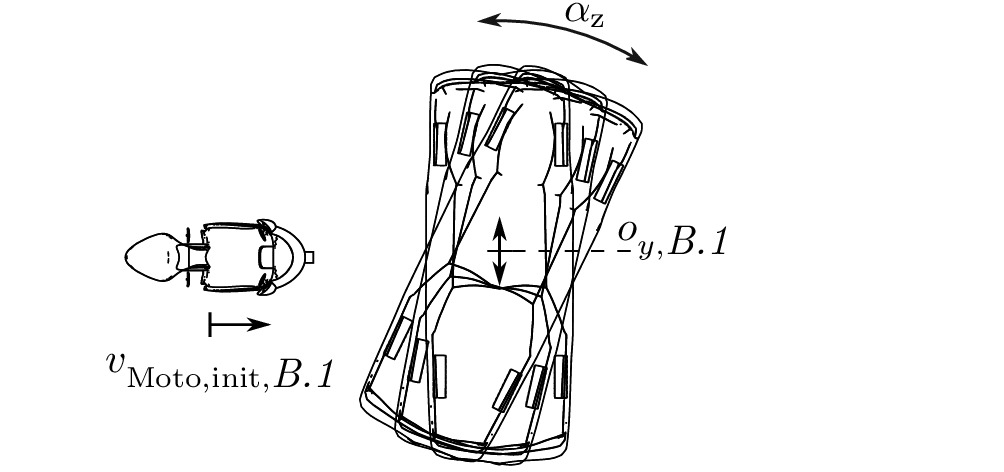}              
  \end{center}
  \caption{Schematic illustration of crash scenario set~\textit{B.1}.}
  \label{fig:b1}
\end{figure}

Set~\textit{B.1} consists of~100 simulations in which the motorcycle with an initial velocity~$v_{\mathrm{Moto,init,B1}}$ approaches a stationary car that is both rotationally~($\alpha_{\mathrm{B1}}$) and laterally~($o_{y,\mathrm{B1}}$) displaced. Figure~\ref{fig:b1} illustrates the structure of the scenario set~\textit{B.1}. By offsetting the car by~$y_{\mathrm{offset}}$, grazing-accidents are included in the scenario database. Table~\ref{tab:b1} shows the range in which the parameters are varied. 

\begin{table}
  \caption{Parameter space for crash scenario set~\textit{B.1}.}
  \label{tab:b1}
  \begin{center} 
	\begin{tabular}[c]{p{3cm} r c r  s[per-mode=symbol]}
	\hline
		parameter & \multicolumn{3}{c}{range} & unit\\ \hline
		$v_{\mathrm{Moto,init,\textit{B.1}}}$ &6.7  &:& 13.4& \si{\m/\s} \\
		$\alpha_{z}$  &0  &:&360&\si{\degree} \\
		$o_{y,\mathrm{\textit{B.1}}}$   &-1.2   &:& 1.2& \si{\m} \\
		\hline
	\end{tabular}
  \end{center}
\end{table}

\subsection*{Scenario Set \textit{B.2}}

\begin{figure}
  \begin{center}
	\includegraphics[width=8.5cm]{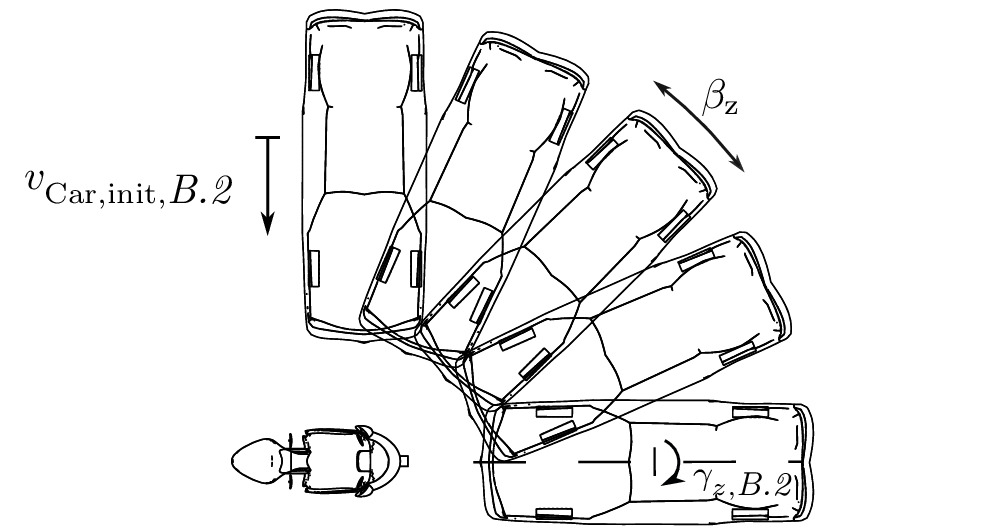}         
    \end{center}
  \caption{Schematic illustration of crash scenario set~\textit{B.2}.}
  \label{fig:b2}
\end{figure}

For more variation, in~\textit{B.2} the setup is inverted with the motorcycle now stationary and the car pointing at the motorcycle at different angles, see Fig.~\ref{fig:b2}. Additionally, the direction of rotation of the car is shifted by~$\gamma_{\mathrm{B2}}$ for further variation of the point of impact.Table~\ref{tab:b2} lists the range of variation for each parameter assigned to set \textit{B.2}.

\begin{table}
  \caption{Parameter space for crash scenario set~\textit{B.2}. }
  \label{tab:b2}
  \begin{center}
	\begin{tabular}[c]{p{3cm}  r c  r s[per-mode=symbol]}
	\hline
			parameter & \multicolumn{3}{c}{range} & unit \\ \hline
			$v_{\mathrm{Car,init,\textit{B.2}}}$ & 6.7  &:& 13.4&  \si{\m/\s} \\
			$\beta_{z}$  & 0   &:&  360&\si{\degree} \\
			$\gamma_{z, \mathrm{\textit{B.2}}}$ & -5   &:& 5&\si{\degree} \\
			\hline
     \end{tabular}
  \end{center}
\end{table}

\subsection*{Validation Set \textit{B.3}}

\begin{figure}
  \begin{center}  
	\includegraphics[width=\textwidth]{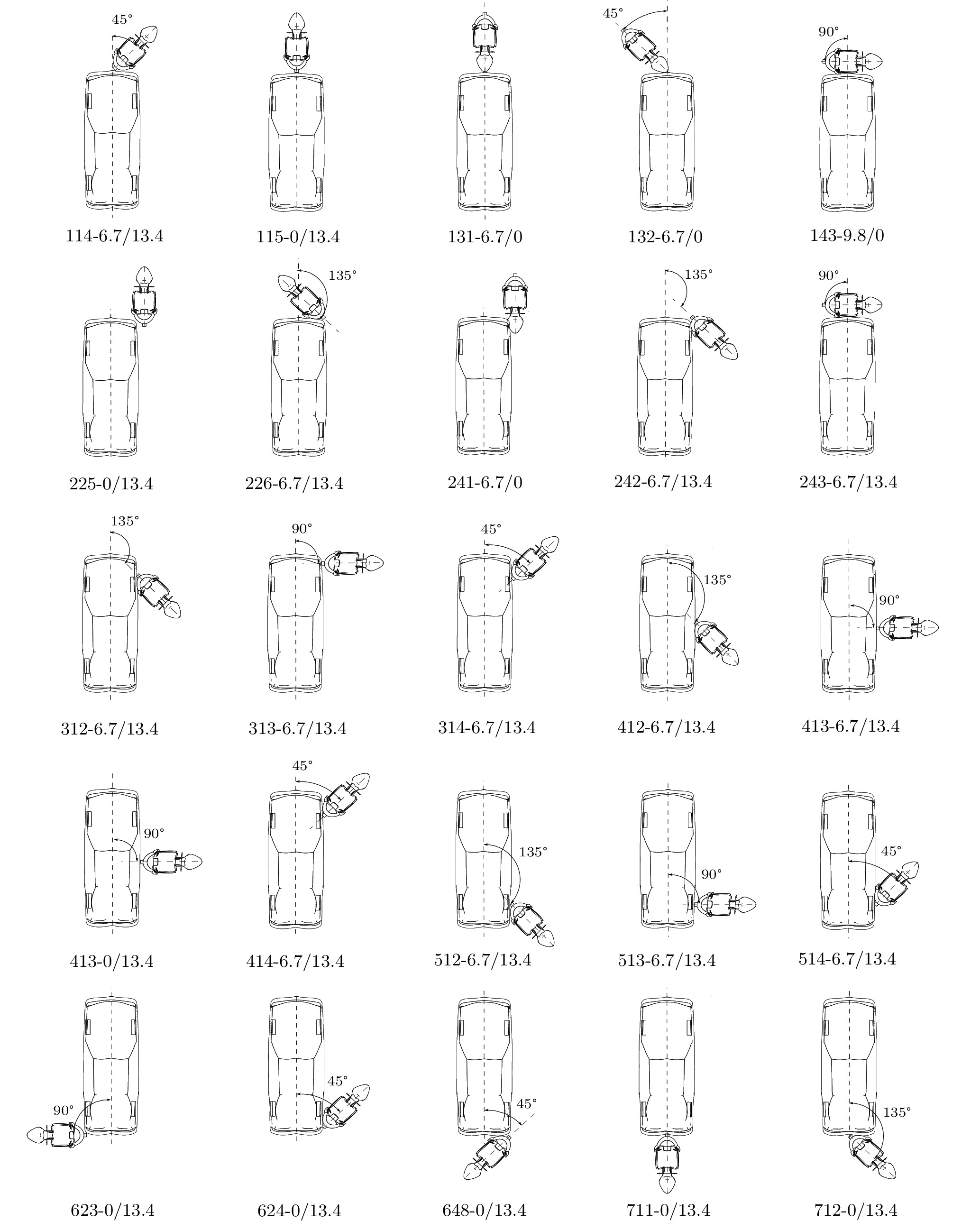} 
  \end{center}
  \caption{Scenarios of the validation set~\textit{B.3} from~\cite{ISO13232}. Each scenario is described by a code~XXX-Y/Z, where~XXX encodes the relative position of car and motorcycle,~Y is the car's, and~Z is the motorcycle's impact velocity in \si[]{\metre\per\s}.}
  \label{fig:b3}
\end{figure}

Validation scenarios are implemented in order to evaluate the final performance of the trained models and, thus, their ability to generalize. Scenarios according to~\cite{ISO13232} are selected for this purpose in order to be representative. The ISO~norm~13232 provides a set of impact configurations based on a statistical analysis of real-world crash events. The full set consists of~25 impact configurations, shown in Fig.~\ref{fig:b3}.

\subsection{Machine Learning Classification}
The remainder of the chapter is devoted to the task of building machine learning models using the \texttt{scikit-learn} toolbox in Python (\cite{scikit-learn}). The section is structured into preprocessing of the raw data, model preselection, and hyperparameter-tuning using grid-search in combination with cross-validation (CV).

\subsubsection{Preprocessing}
Data preprocessing can often lead to a significant performance enhancement compared to working with raw data. Preprocessing includes distribution management, standardization, checking for, and handling of missing values. 

As some classification algorithms tend to perform better or even depend upon standardization, a scaling routine is implemented prior to the classification. Whereas some classification algorithms like decision trees and ensembles do generally not require data standardization, others, like, for example, neural networks, are strongly reliant on it. Otherwise, features that have different magnitudes are not treated equally. Standardization scales each feature to unit variance. In order to be consistent, a transformation vector is computed on the training set and applied to the test set rather than scaling each dataset to unit variance in its own right. The standardized value~$x_{i,m}'$ for each sample~$x_{i,m}$ of timestep~$i$ and feature~$m$ is carried out via 

\begin{equation*}
	x_{i,m}' = \frac{x_{i,m} - \bar{x}_m}{\sigma_{x,m}} 
\end{equation*}

with~$\bar{x}_m$ being the arithmetic mean of feature~$\bm{x}_m$~and $\sigma_{x,m}$ its standard deviation~(\cite{shanker1996}). 

An additional preprocessing method is feature extraction via principal-component-analysis~(PCA). It aims to derive meaningful and non-redundant variables from the original dataset by projecting it to a lower-dimensional space by means of singular value decomposition~(SVD). By this the original dimension of the feature space dimension, which is with a total of~23 features fairly high, could be reduced, This could be beneficial considering that the concluding model is to be run in real-time on an embedded system.
However, the desired effect does not materialize well and the method proves to be ineffective for this application. 
Consequently, it will not be addressed further in this report. For a thorough introduction to PCA and its potential enhancements be referred to~\cite{abdi2010}. 

\subsubsection{Training-Test-Split}
A decisive circumstance to consider when subdividing time-dependent data is that neighboring samples have a tendency to be located in close proximity to each other. A random training test split, like it is often performed on non-time-dependent data, is, henceforth, not an appropriate splitting method~\cite{roberts2017}. 

In the case of this application, training-test-split is carried out on whole coherent simulations itself rather than on samples. As this is a safety-critical application, a~50-50 split is performed, meaning half of the available data is retained for testing to cover a broad range of scenarios. The remaining half is dedicated towards training the models. Additionally, none of the~ISO scenarios of set~\textit{B.3} shall be included in the training set as it is of particular interest to assess how well-trained models generalize on these representative scenarios. 

When training a model for classification purposes, close attention must be paid towards the distribution of the two classes. Since crash-labeled samples are, in this case, much less frequently represented in the raw dataset, non-crash-labeled data requires subsampling in order to achieve equal distribution. The desired distribution is achieved with a sampling rate of twelve. The resulting sizes and distributions of the two datasets are displayed in Tab.~\ref{tab:dataset}.

\begin{table}[htbp]
	\caption{Parameter space for crash scenarios set~\textit{B.2}.}
	\label{tab:dataset}
	 \begin{center}
	\begin{tabular}{p{1.5cm} p{0.15cm} p{2cm}  S[round-mode=places,table-number-alignment=right]}
	\hline
	dataset & & class &  \text{samples}\\ \hline
		&&&\\[-12pt]
		&\hspace{-0.3cm}\rdelim\{{3}{*}[] &non-crash&13842\\
		training & & crash & 10908\\
		& &  total & 24750 \\
		&\hspace{-0.3cm}\rdelim\{{3}{*}[] & non-crash &11238\\
		test& & crash & 11778 \\
		& & total & 23016 \\
		\hline
	\end{tabular}
	\end{center}
   \end{table}

For the final application of the real-time classification model into the virtual "system" of the motorcycle, a sample rate of~\SI[]{2}[]{\kilo\hertz} is selected. This is a sample rate that most commercially available sensors are able to safely handle and which also leaves a sufficient margin of samples for the implementation of an activation threshold. In addition to the above-mentioned training and test datasets, which are no longer subject to a uniform sample rate, individual scenario datasets are prepared, which incorporate the realistic scenarios from~\cite{ISO13232}. Those simulations are synchronized to the selected sample rate of~\SI[]{2}[]{\kilo\hertz} in order to evaluate the model's decisional delay. 

\subsubsection{Model Preselection}
The amount of available classification algorithms is extensive and thus not feasible to investigate exhaustively. A preselection of suitable algorithms is therefore shortlisted. The preselection is designed to incorporate multiple different classification approaches like ensembling, support vector machines~(SVMs), and artificial neural networks~(ANNs) as they are some of the most frequently used algorithms available. Tab. \ref{tab:clfs} lists all preselected models and their underlying method. 

\begin{table}[htbp]
	\caption{Preselection of scikit-learn classification algorithms.}
	\label{tab:clfs}
	 \begin{center}
	\begin{tabular}{lr p{4cm} p{6cm}  l  r}
	\hline
		algorithm & base method\\ \hline
		AdaBoost.M1 & ensembling (boosting)\\
		Gradient Boosting & ensembling (boosting) \\
		Random Forest & ensembling (bagging) \\
		SVM & support vectors \\
		MLP & feedforward neural network \\
		\hline
	\end{tabular}
	\end{center}
   \end{table}

\subsubsection{Hyperparameter-Optimization}
A decisive determinant for the performance of a classification model is the proper choice of its hyperparameters. There are several different approaches to resolve this problem, such as random or gradient searches. In the scope of this paper, grid search parametrization, in combination with cross-validation~(CV), is employed in order to discover the best-fitting parameters for each model. 

Using grid search optimization, the (discrete) hyperparameter space searched within must be predefined. A classification model is trained on each combination of parameters, and, using cross-validation, a mean score is computed for each iteration. The type of score as well as the number of cross-validation folds are user-defined. For this application 20-fold CV in combination with the~$F_1$ score 

\begin{equation*}
	F_1 = 2\:\frac{\mathrm{precision}\cdot \mathrm{recall}}{\mathrm{precision}+\mathrm{recall}} 
\end{equation*}

is selected since the dataset consists itself of a multitude of subsets. The ~$F_1$ is chosen as it combines both precision and recall scores. 

Each parameter's grid is incrementally refined, effectively narrowing the search radius until further refinement no longer yields an improvement. 
Tuned resulting hyperparameters for all five models are listed in Tab. \ref{tab:params}. Only parameters that deviate from the standard parameters are listed. 

\begin{table}[htbp]
	\caption{Tuned hyperparameters for the five classification methods.}
	\label{tab:params}
	 \begin{center}
	\begin{tabular}{p{2.8cm} p{0.5cm} p{4.8cm}  r p{2.5cm}}
	\toprule
		algorithm & & hyperparameter & value\\ 
		\midrule
		&&&\\[-12pt]
		\multirow{4}{*}{AdaBoost} & \hspace{-0.1cm}\rdelim\{{4}{*}[] &number of learners& 80 \\
		& & max. leaner depth & 2\\
		& &  learning rate & 0.6 \\
		&& algorithm & SAMME \\[.2cm]
		
		\multirow{4}{*}{Gradient Boost} & \hspace{-0.1cm}\rdelim\{{4}{*}[] &number of learners& 80 \\
		& & max. leaner depth & 10\\
		& &  learning rate & 0.5 \\
		&& loss function & exponential \\[.2cm]

		\multirow{3}{*}{Random Forest} & \hspace{-0.1cm}\rdelim\{{3}{*}[] &number of learners& 45 \\
		& & max. leaner depth & 17\\
		&& splitting criterion & entropy \\[.2cm]
		
		\multirow{2}{*}{SVM}& \hspace{-0.1cm}\rdelim\{{2}{*}[] &C& 120 \\
		& & gamma & auto\\[.2cm]
		
		\multirow{5}{*}{Neural Net} & \hspace{-0.1cm}\rdelim\{{5}{*}[] &number of hidden layers& 1 \\
		&& hidden layer size & 220\\
		&& initial learning rate & $0.002$ \\
		&& optimization tolerance & $0.0003$ \\
		&& early stopping & enabled \\
		\bottomrule
	\end{tabular}
	\end{center}
   \end{table}

\subsubsection{Baseline Model}
In order to put the performance of the classification models into perspective, a simple threshold based model acts as a baseline. The model is implemented according to \cite{KobayashiMakabe13}, where a set of accelerometers is placed at the fork legs of the motorcycle. The signals are then used to determine if the motorcycle is about to collide. Since the two acceleration signals are averaged, in this paper there is only one accelerometer placed at the front hub. The models only parameter is the threshold at which it detects an impending collision. This threshold is tuned on the training data in the exact same way as for the other models, with its hyperparameter-space being only one dimensional.  

\section{Results and Discussion}
\label{chapter:findings}
In this section, the results from the previously introduced trained classification models are compared to each other and are individually evaluated for fitness.~ML classification-specific criteria on one side, as well as several application-specific criteria, act as a basis for comparison. Furthermore, available sensor data is examined and ranked according to its individual contribution. The computation of feature importance can be helpful to point out wether the feature space dimension can be reduced. The final section describes the individual feature contribution, giving an overview of sensor significance that can assist future work. 

\subsection{Capability Assessment}
The following sections serve as an illustration of the trained classification models' performance. This allows to draw a comparison between the models and assess the level of proficiency that can be expected from a specific model. In addition to a selected set of machine learning performance metrics, a few domain-specific criteria are established. The main objective is to outline the overall performance as broadly as possible in order to make a reasoned choice when selecting a model.   

\subsubsection{Performance Measures}
Performance metrics that are considered to evaluate and compare the achieved training results are the receiver operating characteristic (ROC) curve, explained e.g. in \cite{bishop2006}, which shows the true positive (TP)-false positive (FP) trade-off of a model. The area under the ROC curve~{(AUC)} quantifies the individual trends. Additionally, scores that are computed from the confusion matrix are listed. For the sake of clarity, the resulting confusion matrix of all models is shown in the~appendix (Fig.~\ref{fig:confAda}).

\label{sec:performance}
\subsubsection*{Receiver-Operator Characteristic}

\begin{figure}
    \begin{center}
    \includegraphics[width=14cm]{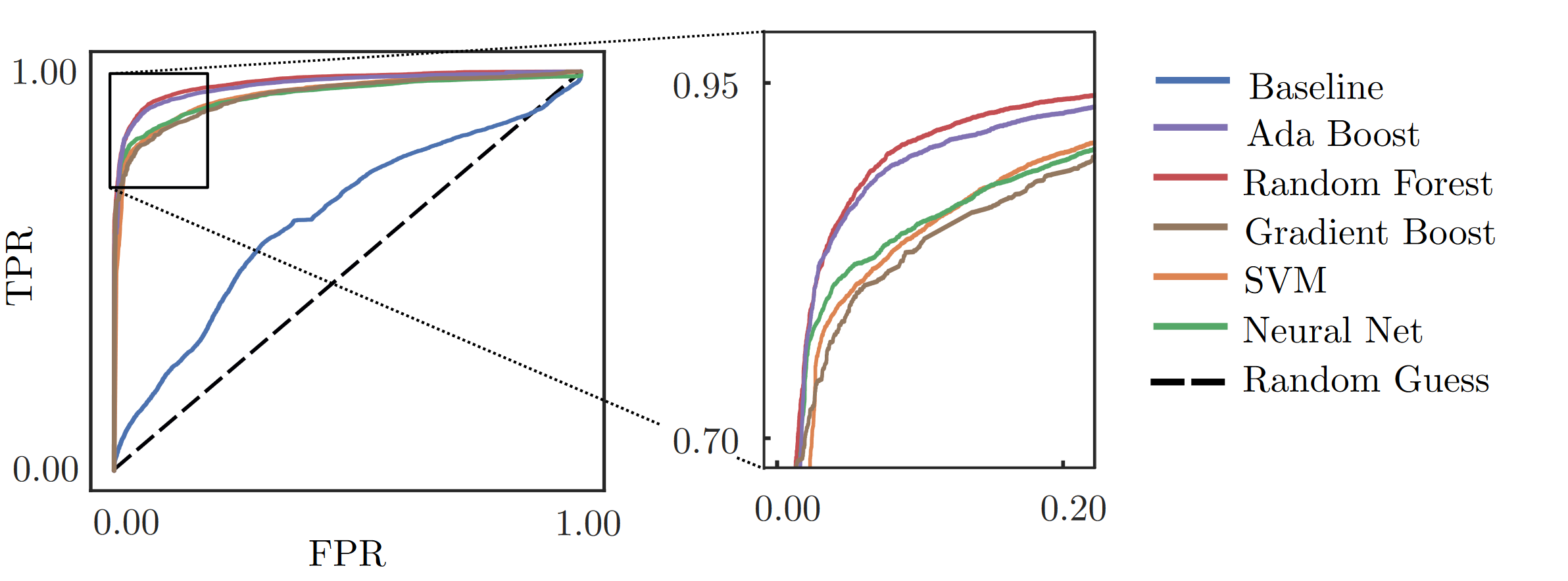}    
    \end{center}
    \caption{Receiver-operator characteristic of all five models. A perfect classification model would reside in the top-left corner, reaching a 100\% TPR while maintaining a 0\% FPR, essentially, mapping each sample flawlessly to its associated class.}
    \label{fig:rocTrained}
\end{figure}

The resulting receiver operator characteristics for all trained models are displayed in Fig. \ref{fig:rocTrained}. All curves lead through the sweet spot in the upper left corner, where classification yields a high TPR while maintaining a low FPR. The results indicate that there is no direct trade-off for all models between achieving a high TP rate~(TPR) and keeping the FP rate~(FPR) low. The ROC curves show an almost identical trend for four of the models, with only the AdaBoost model's performance being marginally poorer. The ROC curve does, therefore, not yet permit any statement about performance fluctuation between the models. 

\subsubsection*{Machine Learning Scores}
\begin{figure}
    \begin{center}
    \includegraphics[width=\textwidth]{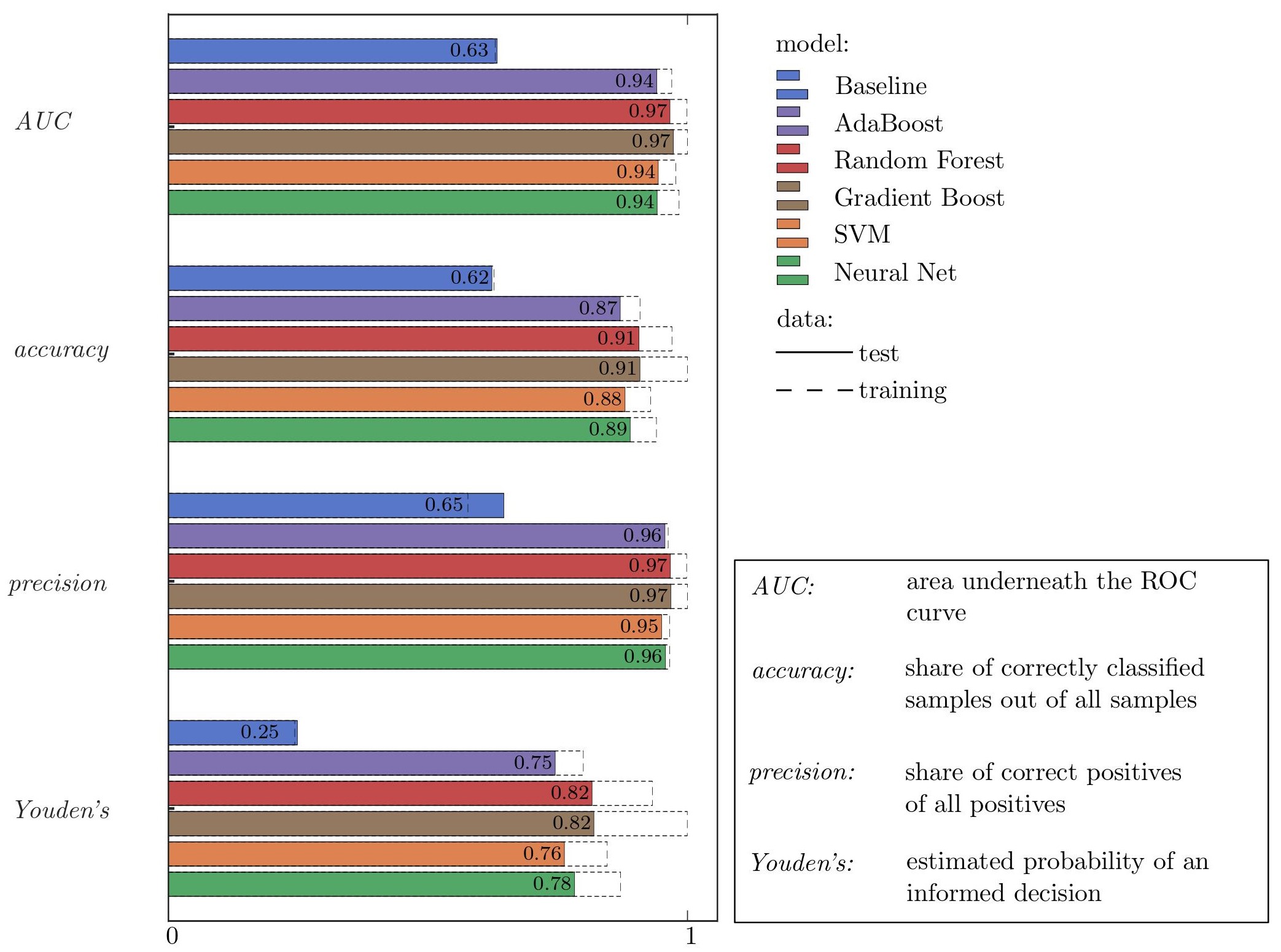}       
    \end{center}
    \caption{Performance scores of all five trained models computed on both training and test data. A large deviation between training and test score is a strong indicator for and overfitted model.}
    \label{fig:scores}
\end{figure}

Since one single index is not able to sufficiently describe the performance of a classification model, a selection of metrics is chosen. The intention is not only to point out which models perform well and which are rendered unfit but also to indicate potential overfitting by computing the score on the training data. Consequently, a well-fitted model will tend to yield similar scores on both sets sacrificing training~\textit{accuracy} for better generalization ability. In contrast, an overfitted model tends to achieve a higher score on the training set than on the test set. The resulting scores are collected in Fig.~\ref{fig:scores}.

\begin{itemize}
\item The \textit{AUC} Score quantifies the trend given by the receiver-operating characteristic, by calculating the area underneath the curve. A higher score means the demand for a better TPR does not tend to sacrifice a models FPR and vice versa.
\item \textit{Accuracy} measures the rate of correctly classified samples of both classes out of all samples. It is thus a valuable indicator of overfitting, if a model yields a sufficiently lower score on the test set than on the training set. 
\item The~\textit{precision} score, which in this context accounts for the classifier's ability to not falsely misclassify a non-crash sample as a crash. Sufficient scoring on this assessment is of elementary importance for applying the method presented in this report. Frequently misclassifying class~\textit{A} samples and, thus, falsely initiating deployment of passive safety mechanisms render the implementation of a ML-based crash detection algorithm possibly more harmful than profitable.
\item As a combination of sensitivity and recall, \textit{Youden's} index measures the overall informedness of a model's decision-making process.
\end{itemize}

Summarizing the scores and discrepancies presented in Fig.~\ref{fig:scores} permits to conclude the models' performance and generalizing ability. At first glance it is clearly visible, that the baseline model is not able to keep up with the ML models in terms of performance. Its scores are significantly lower than those of the other models. The poor scores indicate, that the model may be able to make a decision, albeit not an informed one.


In comparison to the baseline model the ML models tend to achieve a similar performance, with the \textit{AUC} scores only ranging from $0.94$ to $0.97$ and \textit{precision} being almost identical. The \textit{accuracy} and \textit{Youden's} scores however hold more information. Firstly, the test scores of the Random Forest and Gradient Boost models are slightly higher than that of the other models. However, as the dashed line suggests, their ability to generalize will be impaired due to overfitting. To summarize, all of the five models achieve a similar performance with two of the models being subject to overfitting.


\subsubsection{Crash Prediction Requirements}
Besides the aforementioned performance criteria which assess the classification models themselves, there are certain requirements that are given by the very nature of what is intended to follow the crash prediction algorithm. 
Theses requirements are that~(i)~no false detection is raised when not at risk for accidents, that~(ii)~detection delay is sufficiently short, for the airbag to deploy fully before the rider impacts the motorcycle, and that~(iii)~the model is computationally efficient. 
The first domain requirement results from the fact that the algorithm is intended to initiate passive safety precautions, such as the deployment of the airbag and fastening of the thigh belts. Some of the used airbags are non-deflating and, thus, stay inflated for at least a few seconds. False deployment should, therefore, be avoided at all costs due to the obstruction of visibility and maneuverability as well as rider shock and product reputation. Requirement~(ii) aims to keep the prediction delay of the classification models modest since the airbag inflation takes a comparably long time due to a much larger volume than that of a passenger car. In this context, a prediction time lower than~\SI[]{12}[]{\milli\s} has proven to be sufficient in order for the airbag to inflate fully in time~\cite{engel1992}. Furthermore, the fundamental idea behind this investigation is that the classification model operates in real-time on the motorcycle's embedded system. It is, therefore, reasonable to analyze and compare the latency of the classification models since CPU capacity can be assumed to be seriously limited. 
The requirements are tested on the ISO scenarios presented in Fig.~\ref{fig:b3} as well as a share of sets~\textit{A.1} and~\textit{A.2} that are preserved for the test set and that act as control scenarios. Thus, all data the classifier is tested upon is not included in the training set and is not used for hyperparameter tuning. 

\subsubsection*{Decisional Delay}

Requirement~(i)~and~(ii)~correlate strongly, as can be concluded from a conceptual experiment concerning the ROC curve. If the activation threshold~$n_{\mathrm{activation}}$ is set arbitrarily high, the rate of false positives will diminish as would true positives resulting in the lower-left corner in the ROC plane. In the opposing case that the detection threshold is set arbitrarily low. In this case, no positive value goes undetected but at the cost of misclassifying each negative sample in doing so.  

\begin{figure}
    \begin{center}
    \includegraphics[width=13cm]{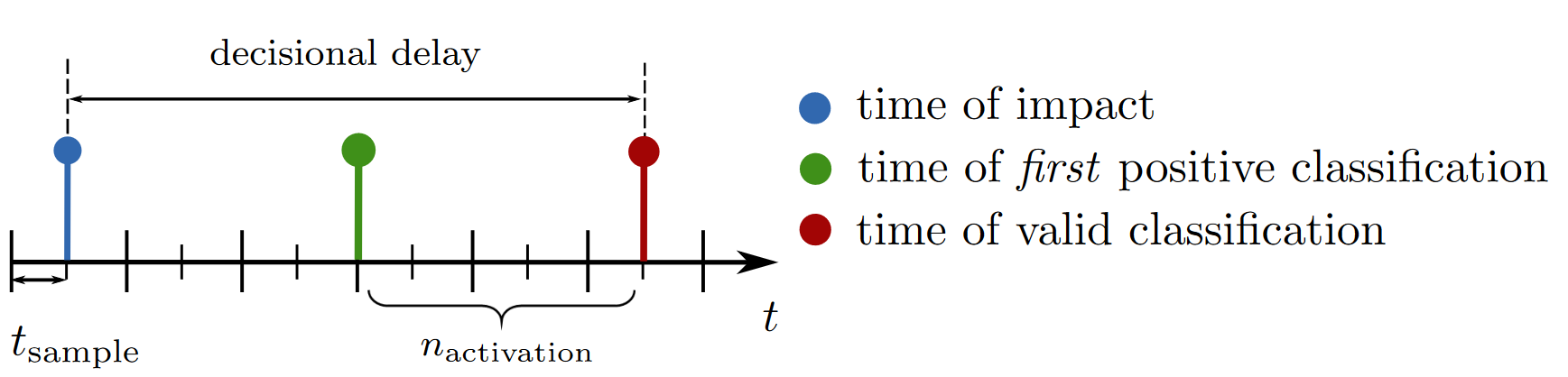}   
    \end{center}
    \caption{An exemplary detection process, beginning at the time of impact and ending with a valid classification. A valid prediction requires a specified number of consecutive positive classifications in order to filter out any false positives.}
    \label{fig:latency}
\end{figure}

Thus, the two requirements are accounted for in the same process. The activation threshold~$n_{\mathrm{activation}}$ assigns a limit to each classification model that tells how many successive samples must be positively classified in order for a valid prediction to be made. The schematic process is explained in Fig.~\ref{fig:latency}. The threshold values are tuned individually for each model so that no false prediction is made in control scenarios. Classification models that are subject to a higher FP rate are consequently assigned a higher threshold~$n_{\mathrm{activation}}$. The individual activation thresholds are listed in Tab.~\ref{tab:thresh}. 

\begin{table}
	\caption{Number of successive samples and equivalent time for a valid classification for each trained classification model.}
	\label{tab:thresh}
	 \begin{center}
	\begin{tabular}[c]{p{2.8cm}  S[table-number-alignment=right] S[table-number-alignment=right] p{1cm}}
	\hline
		model & \text{$n_{\mathrm{activation}}$ } & \text{activation time} & unit\\ \hline
		Baseline &  36 & 18.0 & ms\\
		AdaBoost &  26 & 13.0 & ms\\
		Random Forest &  19 & 9.5 &ms\\
		Gradient Boost &  25 & 12.5& ms\\
		SVM &  16 & 8 & ms\\
		Neural Net &  9 & 4.5 & ms\\

		\hline
	\end{tabular}
	\end{center}
\end{table}

\begin{figure}
    \begin{center}
    \includegraphics[width=14cm]{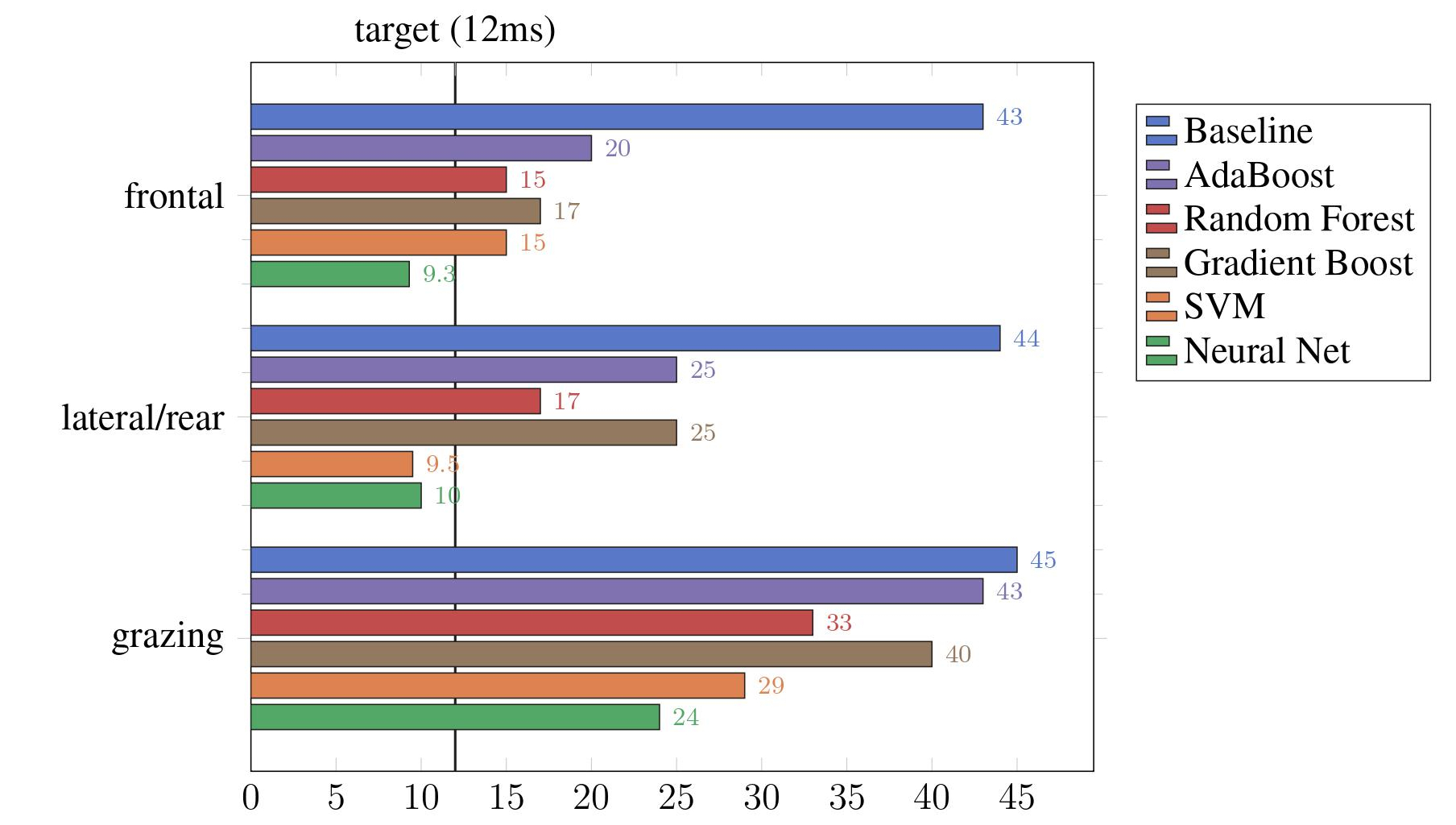}   
    \caption{Prediction delay for the~\cite{ISO13232} control scenarios categorized into 3 accident groups. }
    \label{fig:delay}
\end{center}
\end{figure}

From Tab.~\ref{tab:thresh} it becomes apparent that at a sample rate~$f_{\mathrm{sample}}$ of~\SI[]{2}[]{\kilo\hertz} and an activation threshold~$n_{\mathrm{activation}}=26$ and $25$, the AdaBoost and Gradient Boost models are rendered unfit as they could not meet the required decisional delay of~\SI[]{12}[]{\milli\s} even with a perfect \textit{accuracy} score. The set of ISO crash scenarios is categorized for the purpose of the decision delay assessment. Categories are \textit{frontal} contact crashes, \textit{lateral} or \textit{rear} contact crashes, and lastly, \textit{grazing} accidents. Decision delay is interpreted as the time between initial contact and the moment when $n_{\mathrm{activation}}$ successive positive classifications are reached. The mean decision delay each model achieves in each of the three crash categories is presented in Fig. \ref{fig:delay}. The illustration presents, first of all, an intuitive trend regarding the three crash categories. The delay when predicting \textit{frontal} accidents, on average, is significantly lower than that of \textit{grazing} accidents, meaning they are effectively easier to predict. This can partly be attributed to the acceleration that the motorcycle exhibits during an accident, which is generally larger in magnitude in \textit{frontal} crashes than during \textit{grazing}. The figure shows that only the Neural Net model is capable of making an informed decision within the time limit for \textit{frontal}, \textit{lateral}, and \textit{rear-end} accidents. The SVM model, though failing at detecting \textit{frontal} accidents, predicts \textit{lateral} and \textit{rear-end} accidents in due time. The other models are not able to decide within the required period of time as their decisional threshold does not allow for as much agility. \textit{Grazing} accidents pose a serious challenge for all of the models equally, as no model manages to stay within the time limit.


\subsubsection*{Runtime}
To conclude the capability-assessment-investigation, the runtime of each model is evaluated. Runtime, in the context of real-time capability, is the time that elapses between the input of an instruction and the results output by the computing unit. It shall be stated that the results of this evaluation have no context outside this work and are merely a concept to compare the latency distribution of the six trained models. runtime measurement is carried out on an Intel\textsuperscript{\textregistered} Core\texttrademark  i7-2600 CPU which operates at a base frequency of \SI[]{3.4}[]{\giga\hertz}. The runtime mean and standard deviation are computed for each model on the same test set and on the same processing unit. 

The results are listed in Tab.~\ref{tab:latency} for each of the models. The distribution of the mean values shows that incremental sample classification for the Gradient Boost, SVM, and Neural Net model takes considerably less time than it takes for the remaining two models. With latency ranging from \SIrange[range-units=brackets]{0.05}{0.2}{\milli\s} these classifiers require only a fraction of the time of AdaBoost and Random Forest, where latency is distributed from \SIrange[range-units=brackets]{4.29}{5.60}{\milli\s}. It is to be expected that the runtime of the baseline model is small, as its computational cost is moderate. The low runtime of the models Gradient Boost, SVM, and Neural Net compared to that of the remaining models indicate that they should be preferred especially when real-time capability is required.

  \begin{table}
	\caption{Mean computation time of the different models and standard deviation carried out on an Intel\textsuperscript{\textregistered} Core\texttrademark  i7-2600 CPU.}
	\label{tab:latency}
	 \begin{center}
	\begin{tabular}[c]{p{2.8cm}  S[table-number-alignment=right] S[table-number-alignment=right] p{1cm}}
	\hline
		model & \text{runtime mean} & \text{std. dev.} & unit\\ \hline
		Baseline & 0.02& 0.001 & ms\\
		AdaBoost & 4.29  & 0.14 & ms\\
		Random Forest &  5.60 & 2.27 &ms\\
		Gradient Boost &  0.16 & 0.02& ms\\
		SVM &  0.20& 0.004 & ms\\
		Neural Net &  0.05 & 0.002 & ms\\

		\hline
	\end{tabular}
	\end{center}
\end{table}

\subsection{Feature Importance}

\begin{figure}

\begin{center}
    \includegraphics[width=13cm]{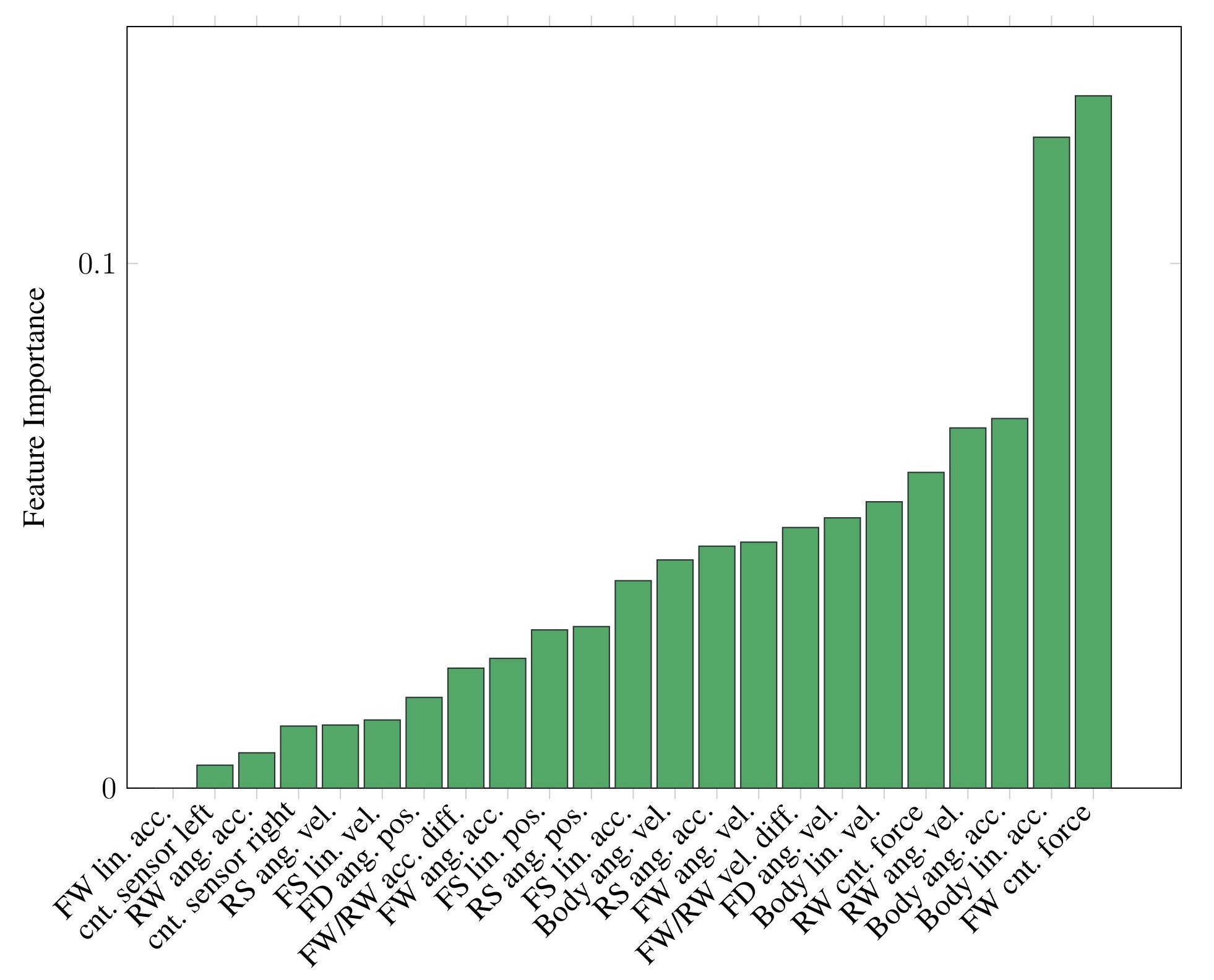}   
    \end{center}
    \caption{Permutation feature importance for the Neural Net model.}
    \label{fig:importance}
\end{figure}

In order to achieve the results shown above, not all features are of equal importance to the classification model. Some features are of greater use in order to differentiate between categories than others. From an economic perspective, it is particularly interesting which features contribute only marginally towards the overall result as it allows the sensors to be dispensed with while maintaining comparable performance. For the following investigation only the results of the MLP are used as it yields the best overall performance and meets the requirements. 
To find the most important features for ANNs, permutation-based feature importance is calculated by performing a number~$k$ of sample interchanges in individual features~$c$ and computes the resulting~\textit{accuracy} score of the corrupted dataset. For each iteration~$k \in \{1,...,n\}$ samples of column~$c$ are interchanged and the resulting score~$s_{k,c}$ is evaluated. A final measure~$i_c$ for the importance of feature~$c$ is given by 

\begin{equation*}
    i_c = s - \frac{1}{n} \sum_{k=1}^n s_{k,c},
\end{equation*}

where~$s$ marks the score on the uncorrupted dataset~\cite{breiman2001}. The contribution of individual features is presented in Fig. \ref{fig:importance} in progressive order so that in the case that feature dimension is limited, the features to omit are highlighted. 


The permutation feature importance ranks that of the motorcycle body's linear acceleration~($\ddot{x}_{\mathrm{Moto}}$) highest, which intuitively appears reasonable. After all, a common condition that all crash scenarios share is an excessive~(negative) acceleration of the system due to a more or less severe impact. However, as shown by~\cite{kuroe2005}, the frame acceleration sensor alone is not a sufficiently agile crash indicator as it responds with a considerable time delay. The delay is mostly due to the deformation that occurs before the frame is actually decelerated. Furthermore, the model assigns a critical weight to the approximated tire pressure $f_{\mathrm{FW}}$, which is derived from the contact force. Front and rear wheel, as well as frame angular acceleration, are also among the top-ranked signals. Signals that only seem to contribute marginally are the suspensions position and velocity both front~($s_{\mathrm{FS}},\:\dot{s}_{\mathrm{FS}}$) and rear~($\varsigma_{\mathrm{RW}},\:\dot{\varsigma}_{\mathrm{RW}}$). The contact sensors that are attached to both sides~($c_{\mathrm{left}}, :c_{\mathrm{right}}$) of the frame also do not particularly enhance the performance. 

Concluding the feature importance investigation, it can be established that acceleration signals generally are of greater benefit towards the overall decision-making quality. A fact that is particularly convenient, as accelerometers are among the most widely used sensors. In contrast, position and velocity signals do not contribute equally as much. This also serves as a potential explanation for the prolonged mean decisional delay that is observed in grazing accidents for all five models. Since in grazing accidents, the induced (negative) acceleration is not as pronounced as in other types of accidents, the acceleration-dominant classifiers fail to respond in due time. This forfeit, however, is admissible since it implies that severe accidents with more pronounced acceleration are reliably detected.
\section{Summary and Conclusion}

The outline of this elaboration is to implement and evaluate machine learning methods to detect motorcycle collisions early and reliably. Simulation data is used for this purpose as it is easier and more cost-saving to acquire. A further benefit is that it allows close monitoring and, thus, enables the utilization of supervised learning. For this reason, the multi-body simulation model is adjusted to incorporate automated data labeling without requiring manual intervention. An extensive database is, henceforth, produced by means of parametrized scenario generation in order to broadly cover both areas of interest.

The simulation model has some limitations, e.g. it does not allow cornering. Thus, this work cannot illustrate all possible accident scenarios. Instead, this work suggests a workflow which must be extended in the future for a holistic consideration. The signal selection and data processing take into account the influences, e.g. the non-consideration of cornering by not inducing "false knowledge" into the database. The database is used to train a selected set of machine learning classification models. The mindfully preselected algorithms that are under consideration are three decision tree ensemble models, an support vector machine as well as an artificial neural network. In order to have a better understanding of the results, a primitive, threshold based model acts as a baseline for comparison. 

Challenges that are imposed on the prediction model are that~(i)~it raises no false detection during normal operation,~(ii)~it responds sufficiently quick to account for the airbag's prolonged inflation time, and~(iii)~it is computationally efficient for realtime application. In order to evaluate each individual classification model's performance and to rate their fitness, a sophisticated scoring procedure is established in Section~\ref{chapter:findings}, allowing for a reasoned choice. Performance is evaluated on a set of frequent crash scenarios from~ISO~13232 to be representative. The classification models are individually tuned towards not making a potentially harmful false prediction. Only one out of the five trained models is able to reliably identify the more severe crashes flawlessly and in due time, with only the less harmful grazing accidents posing a challenge. Despite being fail-safe during non-crash driving operation, this model predicts crashes with sufficient agility for the airbag to inflate within the prescribed time. At the same time the neural net model proves to be amongst the lowest in terms of computation time. 

Although the scores of other models were superior, the Neural Net model still prevails, as it provides the most reliable classification and thus requires the lowest activation threshold of all the models. Furthermore, this work gives insight into the significance of individual sensor data via feature importance assessment. This examination reveals that, in general, acceleration signals are weighted much more favorably than position and velocity signals. This circumstance renders grazing accidents particularly difficult to predict, as the acceleration imposed on the motorcycle is considerably lower than in other collision types.

Despite the exciting contributions of this study in detecting motorcycle collisions for a variety of simulated scenarios, it is important to acknowledge its limitations, particularly with regards to the restricted computational capacity of in-vehicle control units and the lack of consideration of lateral dynamics due to simulation model shortcomings. While realistic boundary conditions are adopted, wherever feasible (e.g. sensor sample time, ISO crash scenarios), it should be noted that, for the sake of simplification, ideal measurement hardware is assumed. Any loss in signal quality, that any commercially available sensor would impose is neglected, and must be taken into account in future studies. 
Therefore, in subsequent research, it is of interest to first consider even more realistic and extensive scenarios and mimic real sensor measurements by adding noise to the data or using sensor models. In a second step, it is of interest to investigate the suitability of the prediction algorithms for real-time use on weak hardware.



\section*{Acknowledgements}

Funded by Deutsche Forschungsgemeinschaft (DFG, German Research Foundation) under Germany's Excellence Strategy - EXC 2075 - 390740016. We acknowledge the support by the Stuttgart Center for Simulation Science (SimTech). All authors would like to thank the Ministry of Science, Research, and Arts of the Federal State of Baden‐W{\"u}rttemberg for the financial support of the project "Transportable echtzeitf{\"a}hige digitale Zwillinge~(TEDZ) -- f{\"u}r den sicheren und komfortablen Betrieb neuer Mobilit{\"a}tskonzepte" within the "InnovationsCampus Mobilit{\"a}t der Zukunft".

\section*{Data Availability Statement}
The datasets generated for this study can be found in the DaRUS repository~\cite{darus-3301}. The dataset provides time histories of sensor data in *.csv format, classified into "accident" and "no accident".

\printbibliography

%
%
\begin{appendix}
\section*{Appendix}
\begin{figure}[htp]
    \begin{center}  
    \includegraphics[]{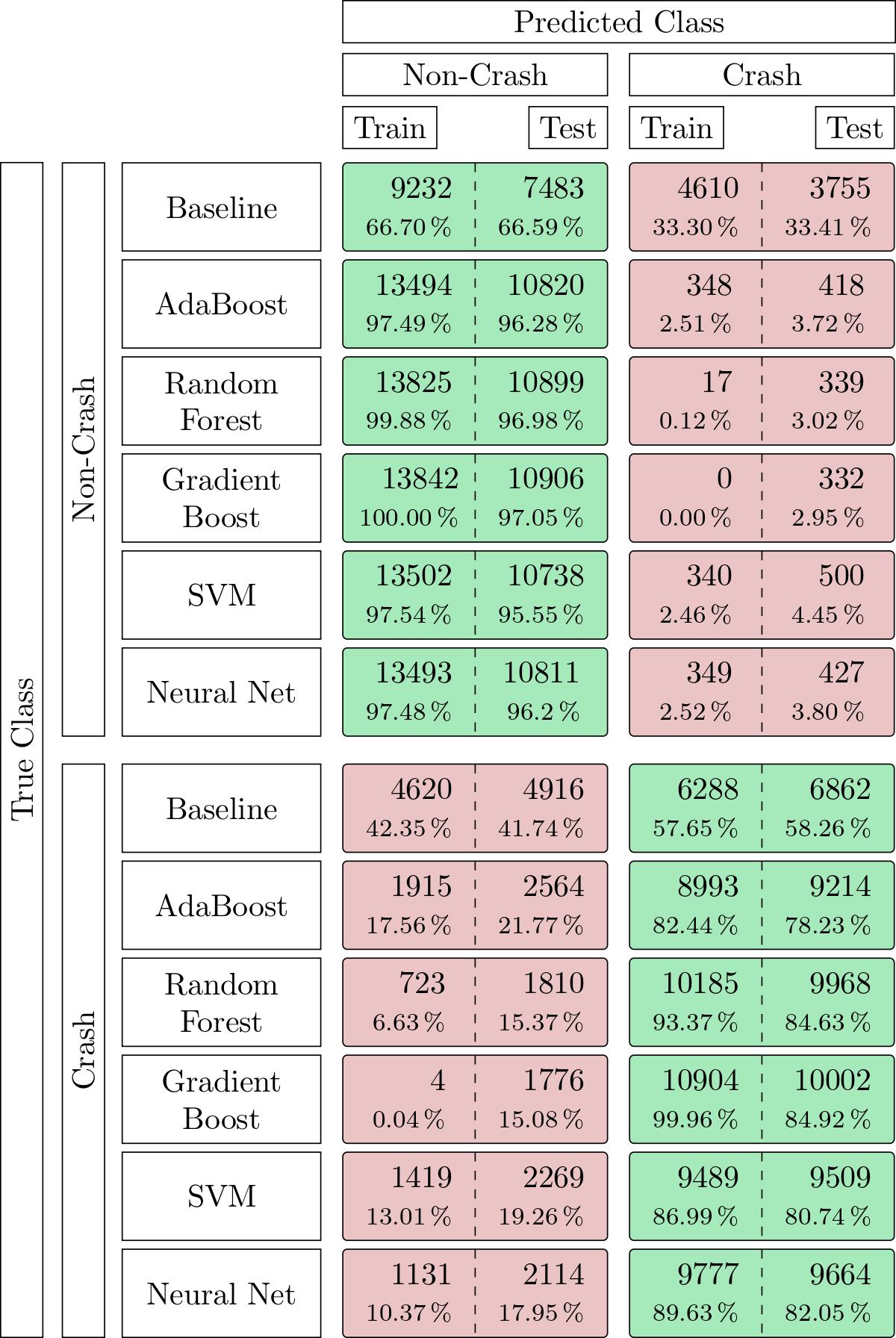} 
    \end{center}
    \caption{Confusion matrix of all classification models evaluated on both training and test data.}
    \label{fig:confAda}
    \end{figure}
\end{appendix}

\end{document}